\documentclass[lettersize,journal]{IEEEtran}
\usepackage{amsmath,amsfonts}
\usepackage{algorithmic}
\usepackage{algorithm}
\usepackage{array}
\usepackage[caption=false,font=normalsize,labelfont=sf,textfont=sf]{subfig}
\usepackage{textcomp}
\usepackage{stfloats}
\usepackage{url}
\usepackage{verbatim}
\usepackage{graphicx}
\usepackage{cite}
\usepackage{booktabs}
\begin{document}

\title{Image Generation Based on Image Style Extraction}

\author{Shuochen Chang}



\maketitle

\begin{abstract}
image generation based on text-to-image generation models is a task with practical application scenarios. The challenge remains that fine-grained styles cannot be precisely described and controlled in natural language, while the guidance information of stylized reference images is difficult to be directly aligned with the textual conditions of traditional textual guidance generation. Among the training-based method, some related works implement the reference image as a guidance condition by introducing a new cross-attention mechanism in the denoising network, while some others simply map the stylized image to the text space as a generative guidance. The main problem of the above works is that the content of the stylized reference image is coupled with the stylistic information, which leads to mutual interference between the image and the textual control condition, so that on the one hand, the newly generated image has similarities with the reference image in content, and on the other hand, the semantic information of the textual cues may be lost. This study focuses on how to maximize the generative capability of the pretrained generative model, by obtaining fine-grained stylistic representations from a single given stylistic reference image, and injecting the stylistic representations into the generative body without changing the structural framework of the downstream generative model, so as to achieve fine-grained controlled stylized image generation. In this study, we propose a three-stage training style extraction-based image generation method, which uses a style encoder and a style projection layer to align the style representations with the textual representations to realize fine-grained textual cue-based style guide generation. In addition, this study constructs the Style30k-captions dataset, whose samples contain a triad of images, style labels, and text descriptions, to train the style encoder and style projection layer in this experiment. Evaluation of the experimental results shows that the method in this study is able to extract fine-grained stylistic features of reference images and use them in a text-to-image generation model to generate a brand new image that conforms to the style of the target image and is consistent with the textual instructions.
 
\end{abstract}


\section{Introduction}
\label{sec:intro}

\IEEEPARstart{T}{he} advent of large-scale generative models pre-trained on diffusion principles, exemplified by Stable Diffusion \cite{dm, sd}, has marked a significant milestone in artificial intelligence. The remarkable quality and text-conditional controllability of these models signify a major breakthrough, establishing diffusion models as a dominant force that is progressively supplanting traditional frameworks like Variational Auto-Encoders (VAEs) \cite{vae} and Generative Adversarial Networks (GANs) \cite{gan}. Consequently, Artificial Intelligence Generated Content (AIGC) has emerged as one of the most dynamic and impactful research frontiers in computer vision.

The application landscape of AIGC is vast, encompassing a variety of downstream tasks such as image generation, editing, composition, harmonization, and inpainting. In these applications, diffusion models typically serve as the generative backbone. Through a series of operations—including masking, noise injection, and guided denoising—these models can produce novel images that meet specific requirements, demonstrating high degrees of quality and fidelity.

Among these applications, Text-to-Image (T2I) generation has become exceptionally popular, largely due to advancements in multimodal alignment, particularly the Contrastive Language-Image Pre-Training (CLIP) model \cite{clip}. CLIP enables the alignment of text and images within a shared high-dimensional embedding space, allowing diffusion models to use text as a conditional guide to dictate the semantic content of the generated output. While this has greatly enhanced the utility of AIGC, the generation of images with \textit{fine-grained} and \textit{specific} styles remains a formidable challenge. Artistic styles are often nuanced and complex, making them difficult to articulate precisely with natural language. Furthermore, pre-trained models often struggle with zero-shot generation of styles they have not encountered during training.

Traditional style transfer methods \cite{st1} focused on iteratively optimizing a generated image to match the style loss of a style reference and the content loss of a content reference. However, these methods cannot generate novel content guided by textual prompts. Subsequent approaches based on GANs, such as Conditional GANs (cGANs) \cite{cgan} and CycleGAN \cite{cyclegan}, improved style transfer and stylized generation. Yet, they often fall short in terms of image quality and the granularity of style representation compared to modern diffusion models. Crucially, they lack the intrinsic ability for conditional text guidance, severely limiting their applicability in creating content-specific stylized images.

To overcome these challenges, a growing body of work now focuses on stylized image generation using diffusion models. Prominent solutions include Textual Inversion \cite{ti}, which learns a new pseudo-word in the embedding space to capture a specific concept; DreamBooth \cite{db}, which fine-tunes the entire diffusion model to specialize in a subject or style; and adapter-based methods like IP-Adapter \cite{ip}, which inject image-based conditioning into the cross-attention layers. These methods ingeniously modify different stages of the diffusion pipeline, enabling precise style control by using reference images to circumvent the descriptive limitations of natural language, all while preserving the generative power of the foundational model.

Building on these insights, this paper proposes a novel, systematic training methodology for end-to-end style feature extraction and generation from a single reference image. Our approach is centered around \texttt{Stable Diffusion v1.4}. We begin by constructing a high-quality, style-content decoupled dataset named \texttt{Style30k-captions}, which we create by generating detailed content captions for the style-rich images in the \texttt{Style30k} dataset \cite{oh} using GPT-4o. Our training pipeline is divided into three distinct stages: 
1) \textbf{Stylized Textual Inversion:} We learn a style vector for each image that is aligned with the text embedding space, effectively decoupling style from the textual content description. 
2) \textbf{Style Encoder Pre-training:} We pre-train a CLIP-based vision encoder to map input images to their corresponding style vectors, enhancing the clustering of style features.
3) \textbf{Joint Fine-tuning:} Inspired by the architecture of large multimodal models \cite{qwen, intern}, we fine-tune the style encoder and a new linear projection layer jointly. This stage aligns the extracted visual style features with the text embedding space, enabling precise, controllable, and fine-grained stylized generation.

The primary contributions of this work are as follows:
\begin{itemize}
    \item We introduce \texttt{Style30k-captions}, a large-scale dataset of 30,000 high-quality image-text pairs where fine-grained style is preserved in the image and content is described in the caption, facilitating style-content decoupled learning.
    \item We propose an innovative three-stage training framework that synergistically integrates textual inversion, encoder pre-training, and joint fine-tuning. This framework achieves a robust balance between fine-grained style control and coarse-grained style categorization.
    \item Our method enables end-to-end style extraction and generation from a single reference image, significantly enhancing the fine-grained style fidelity of diffusion models while retaining their inherent content generation capabilities.
    \item The proposed framework offers a new paradigm for conditional image generation that is extensible to other tasks, such as personalized generation and instance-guided image editing, and provides a reference architecture for other generative modalities.
\end{itemize}

\section{Related Work}
\label{sec:related_work}

This section reviews existing algorithms for stylized image generation, primarily focusing on methods built upon diffusion models. We categorize and discuss these approaches based on their primary driving modality: text-driven methods that generate new images from textual prompts, and image-driven methods that transfer style onto existing content images.

\subsection{Text-Driven Stylized Generation}
\label{ssec:text_driven}

Text-driven stylized generation leverages textual prompts to guide a generative model in creating novel images that adhere to a reference style. The primary advantage of this paradigm is its flexibility and its ability to maximally preserve and utilize the powerful prior knowledge embedded within pre-trained text-to-image models. We survey several mainstream approaches below.

\subsubsection{Image Encoder-Based Methods}
With the continuous improvement of multimodal models' visual encoding and image-text alignment capabilities, pre-trained image encoders can provide powerful semantic guidance for diffusion models. These methods typically extract style features from a reference image and use them, either in conjunction with text features or independently, to direct the generation process.

IP-Adapter \cite{ip} introduces a lightweight adapter that uses high-level image features from a CLIP model to guide the diffusion denoising process. It employs a decoupled cross-attention mechanism, ensuring that text and image features control the noise prediction through separate attention layers. This endows the diffusion model with an "image prompt" capability without altering its core architecture, enabling fine-grained style control.

Similarly, ArtAdapter \cite{artadapter} utilizes a pre-trained VGG network, a deep Convolutional Neural Network (CNN), to capture a hierarchy of style representations. By tapping into intermediate outputs from different depths of the network, it extracts features ranging from low-level (shallow layers) to high-level (deep layers). These style representations are then mapped into the text embedding space and integrated via a comparable decoupled cross-attention adapter.

InstantStyle \cite{instantstyle} introduces an innovative yet simple method that leverages the shared semantic space of CLIP. It decouples style and content by subtracting the text feature of the image's content from its corresponding image feature, isolating the style representation. Furthermore, the study found that injecting this decoupled style feature into specific, style-sensitive layers of the diffusion U-Net, rather than throughout the entire network, reduces content leakage and enhances style control.

EasyRef \cite{easyref} leverages the instruction-following capabilities of Multimodal Large Language Models (MLLMs) to enhance controllable generation, particularly for tasks involving multiple reference images. Instead of simply averaging the CLIP embeddings of multiple references, it uses an MLLM to process the reference images and interact with a set of learnable query tokens. The resulting tokens are projected to align with the text guidance space of the diffusion model, enabling plug-and-play customized generation from single or multiple references.

ArtCrafter \cite{ArtCrafter} proposes a method for style feature extraction and mixed-modality fusion. It first extracts style features from a reference image using a CLIP model and a feed-forward network. It then introduces an attention-based fusion mechanism to enhance the integration of these visual style features with the high-level semantic information from text prompts. This fused multimodal embedding allows the diffusion model to more effectively synthesize images that are both semantically precise and stylistically faithful.

\subsubsection{Textual Inversion-Based Methods}
The seminal work of Textual Inversion \cite{ti} first proposed learning a "pseudo-word" to enable personalized generation from text-guided diffusion models. To handle concepts not easily described by natural language, it introduces a learnable word embedding that is optimized by reconstructing a few example images. This lightweight approach allows for the controllable generation of specific instances or styles using the learned embedding in a text prompt.

InST \cite{inst} builds upon textual inversion by integrating both text-driven and image-driven approaches. It refines the training process with a novel attention-based optimization for the word vector, guided by CLIP features. This allows the learned pseudo-word to better incorporate the visual information from the reference style image, achieving more granular style control.

The work "An Image is Worth Multiple Words" \cite{imword} expands the application of textual inversion to multi-concept learning. It addresses the limitation of prior work, which typically learns only a single concept from a reference image. The proposed method uses natural language descriptions to guide the model in learning multiple concepts simultaneously from a single image, enabling precise content control and local editing.

\subsubsection{Attention Feature Swapping Methods}
This line of work manipulates the internal attention mechanisms of diffusion models to achieve style control without explicit training. Cross-image attention \cite{crossimage} focuses on zero-shot appearance transfer. It enables the migration of attributes like shape, color, and texture from a specific instance in one image to a target in another by fusing cross-attention features within the diffusion U-Net, showcasing robust, training-free appearance control.

StyleAligned \cite{stylealign} addresses the challenge of maintaining style consistency across a batch of generated images. By sharing attention features among a set of images during their parallel generation process, often normalized with methods like AdaIN, it ensures that all images in the batch adhere to a highly consistent visual style.

Visual Style Prompting (VSP) \cite{vsp} concentrates on the self-attention features during the denoising process. It proposes a novel self-attention design where the keys and values in the later upsampling layers are replaced with those from a style reference image at specific timesteps. This targeted replacement injects style features while preventing the content of the reference image from leaking into the final output.

\subsubsection{Fine-Tuning-Based Methods}
StyleDrop \cite{styledrop} aims to solve the out-of-distribution problem when generating images with highly specific design aesthetics. It utilizes adapter-based fine-tuning and introduces an iterative feedback training framework, where images sampled from a previously trained adapter are used as new training data to further refine its parameters, progressively improving style-content consistency.

DreamStyler \cite{DreamStyle} also addresses the inadequacy of natural language for describing nuanced artistic styles. It proposes a multi-stage textual inversion algorithm that learns style embeddings in an extended text embedding space. Different style embeddings are applied at different stages of the denoising process, allowing them to adapt to the dynamic changes of the diffusion model and capture style attributes more accurately.

ControlStyle \cite{ControlStyle}, inspired by the design of ControlNet, tackles text-driven stylized generation by adding a parallel control network. This network extracts style features from a reference image and feeds them into the main denoising U-Net via zero-convolution layers, enabling precise modulation of the output style. To prevent content degradation, it also employs style and content regularization during diffusion.

\subsubsection{Training-Free Methods}
Attention Distillation \cite{attndis} transfers the texture and appearance from a given image to a newly generated one via inference-time optimization. It introduces an attention distillation loss, which minimizes the difference between the self-attention scores of the current inference step and those of the style reference image, alongside a content preservation loss to guide the generation without any training.

RB-Modulation \cite{RB-Modulation} tackles the difficulty of style extraction and style-content disentanglement in training-free scenarios. It applies stochastic optimal control to the diffusion process, incorporating style features into the controller to modulate the drift field of the denoising dynamics. Combined with a cross-attention-based feature fusion scheme, this improves the quality and controllability of the generated images.

\subsection{Image-Driven Stylized Generation}
\label{ssec:image_driven}

Image-driven stylized generation, often referred to as image style transfer, primarily aims to apply the style of a reference image onto a target content image. The core challenges in this task are effective style transfer and robust content preservation, making it more about editing and stylizing existing content rather than generating novel scenes from scratch.

\subsubsection{Text Encoding-Based Methods}
DiffStyler \cite{DiffStyler} proposes a novel dual-diffusion architecture for text-driven image stylization. Crucially, instead of starting from Gaussian noise, it initiates the reverse denoising process from a noised version of the content image, which helps preserve its structure. It then employs a combination of a style-specialized diffusion model and a general-purpose one during sampling to strike a balance between style fidelity and content preservation.

ZeCon \cite{Zero-Shot} focuses on text-guided image style transfer without altering the image content. It introduces a zero-shot contrastive loss that operates on the intermediate noise maps of the pre-trained diffusion model. By enforcing consistency between the noised content image and the stylized generated image at corresponding steps, it ensures structural and semantic integrity.

Stylebooth \cite{styleboth} proposes a unified framework for aligning reference style images with text prompts. It uses a fixed text template with a placeholder and learns a linear projection to map the extracted features of a style image into the text embedding space, effectively "filling in the blank." It also adopts the strategy of using a noised content image as the starting point for denoising, ensuring fine-grained style control while maintaining the spatial structure.

\subsubsection{Other Approaches}
Arbitrary Instance Style Transfer \cite{arbit} enables the application of a given style to any specific instance within a content image. The method leverages a pre-trained Mask R-CNN for instance segmentation. It then uses a pre-trained VGG-19 as an image encoder to extract features from both content and style images, performs the stylization, and finally re-integrates the stylized instance back into the original content scene.

\section{A Three-Stage Framework for Fine-Grained Stylized Image Generation}
\label{sec:method}

This chapter details the three-stage training framework we designed to achieve fine-grained stylized image generation. While much of the existing work focuses on improving performance aspects such as style diversity, granularity, content fidelity, and text-to-image consistency, our research introduces a new paradigm. By decoupling the content and style from a reference image, we can extract fine-grained style features and leverage the full capabilities of the pre-trained Stable Diffusion model to exert precise control over the generation process. This chapter will first introduce the dataset used for training and evaluation. Subsequently, we will delve into the design rationale and specific implementation of our three core training stages. 
\begin{itemize}
    \item \textbf{Stage 1} employs textual inversion to iteratively optimize a fine-grained style vector for each image in our dataset, aligning it with Stable Diffusion's text embedding space.
    \item \textbf{Stage 2} shifts the paradigm by pre-training a style encoder and a projection layer, enabling the direct extraction of a style vector from an input image in a single forward pass.
    \item \textbf{Stage 3} involves a joint fine-tuning of the pre-trained style encoder and projection layer to further enhance their performance, directly optimizing for the downstream task of generating high-quality, fine-grained stylized images.
\end{itemize}
Finally, we will describe the complete inference pipeline for controllable stylized image generation using our trained modules.

\subsection{Dataset Construction and Preprocessing}
\label{ssec:dataset}

The quality of our dataset is paramount, as it directly influences the efficacy of each training stage and, ultimately, the final generation quality and style control. Our first step was to construct a suitable dataset for our task.

We selected the open-source \texttt{Style30k} dataset \cite{oh} as our foundation. \texttt{Style30k} comprises 25,868 high-quality, high-resolution images that exhibit a diverse and fine-grained range of artistic styles, categorized into 1,121 distinct classes such as watercolor, classical oil painting, anime, minimalism, and graphic design. Crucially, \texttt{Style30k} provides a coarse-grained semantic style label (e.g., "Impressionism", "2D Design") for each class, which is vital for training a model's perception of style.

While text prompts can effectively describe content, they often fail to capture the nuances of fine-grained style. To address this, our approach decouples content from style, with the style information being provided exclusively by a reference image. To achieve this at the dataset level, we extended \texttt{Style30k} by generating content-only captions for each image. For efficiency and consistency, we employed the powerful GPT-4o model for automatic annotation. The core objective was to generate a text caption $C_i$ for each image $I_i$ that describes its content purely, devoid of any stylistic language. We meticulously engineered the following prompt to guide the model:
\begin{center}
\begin{minipage}{0.9\columnwidth}
\textit{Describe only the content and subject of this image in short words. Must ignore any artistic style and Do not mention the artist, the style. Focus purely on what objects or subjects are depicted. Do not use words like 'abstract', 'colorful', 'abstract expressionism'. Do not mention colors, textures, brushstrokes, lighting style, artistic movement, or overall mood.}
\end{minipage}
\end{center}

After automated annotation, we performed manual sampling and inspection to verify the quality of the generated captions and their adherence to our strict constraints. The results were highly satisfactory, with minimal style information leakage or content misrepresentation. To explicitly separate the content and style slots in our text prompts, we appended a fixed suffix to each caption: \textit{"in the style of [*]."}. Here, \textit{[*]} serves as a placeholder for the style information that will be learned during Stage 1. This process resulted in a new image-text pair dataset, which we name \textbf{Style30k-captions}. It consists of 25,868 $(I, C)$ pairs, where $I$ provides the fine-grained style information and $C$ provides the decoupled content description.

\subsection{Data Preprocessing Pipeline}
\label{ssec:preprocessing}

Before training, each data point $(I, C, Tag)$ from our dataset, where $Tag$ is the original coarse style label, undergoes preprocessing.
For an image $I$, when it serves as input to our CLIP-based style encoder $E_{\text{style}}$ (in Stages 2 and 3), its resolution is resized to $224 \times 224$ pixels, matching CLIP's pre-training configuration. The image is then normalized based on the mean and standard deviation expected by the pre-trained model.

For text data, both the content caption $C$ and the style tag $Tag$ are processed using the tokenizer corresponding to the CLIP text encoder used in Stable Diffusion v1.4. This tokenizer converts text strings into a sequence of token IDs. As the text encoder has a maximum input length of 77 tokens, sequences are either padded or truncated to meet this requirement. After preprocessing, all data is converted into a numerical format suitable for direct use by our models.

\section{Stage 1: Style Vector Learning via Textual Inversion}
\label{sec:stage1}

The objective of Stage 1 is to learn a dedicated vector, termed the style vector $V_{\text{style}}^{(i)}$, for each image $I_i$ in \texttt{Style30k-captions}. This vector is designed to capture the image's fine-grained visual style and reside within the same high-dimensional space as the text embeddings used by Stable Diffusion (SD). We achieve this using Textual Inversion, a technique that finds a corresponding "pseudo-word" embedding within the model's text embedding space for a given visual concept.

To enhance expressive power, we define each style vector $V_{\text{style}}^{(i)}$ as a sequence of 8 tokens. Thus, $V_{\text{style}}^{(i)} \in \mathbb{R}^{8 \times d_{\text{text}}}$, where $d_{\text{text}}$ is the embedding dimension of the SD model's CLIP text encoder ($d_{\text{text}}=768$ for the ViT-L/14 used in SD v1.4). This multi-token design allows for a richer encoding of nuanced style attributes. Each style vector is initialized with values sampled from a Gaussian distribution, $\mathcal{N}(0, 0.02^2)$. During this stage, all original parameters of the SD v1.4 model—including the U-Net noise predictor $\epsilon_{\theta}$, the VAE encoder/decoder ($E_{\text{VAE}}, D_{\text{VAE}}$), and the CLIP text encoder $E_{\text{text}}$—are completely frozen. The optimization is performed solely on the style vector $V_{\text{style}}^{(i)}$ for each image.

For each pair $(I_i, C_i)$ in the training set, we perform an independent optimization. The goal is to find the optimal $V_{\text{style}}^{(i)}$ that, when combined with the content description $C_i$, best reconstructs the original image $I_i$ in the latent space. The process is as follows: we first encode the caption $C_i$ into content embeddings $E_{\text{text}}(C_i)$. We then concatenate the learnable style vector $V_{\text{style}}^{(i)}$ with these content embeddings to form the combined conditioning vector, $Cond = [V_{\text{style}}^{(i)}; E_{\text{text}}(C_i)]$. Concurrently, the image $I_i$ is encoded into its latent representation $z_0 = E_{\text{VAE}}(I_i)$. We simulate the diffusion process by adding noise $\epsilon$ at a random timestep $t$, producing $z_t$. The frozen U-Net $\epsilon_{\theta}$ then predicts the noise from $z_t$ conditioned on $Cond$. The optimization minimizes the mean squared error (MSE) between the predicted noise $\epsilon_{\text{pred}}$ and the ground-truth noise $\epsilon$:
\begin{equation}
\label{eq:recon_loss_stage1}
\mathcal{L}_{\text{recon}}(V_{\text{style}}^{(i)}) = \mathbb{E}_{t, \epsilon} \left\| \epsilon - \epsilon_{\theta}(z_t, t, Cond) \right\|_2^2
\end{equation}
The gradient of this loss is used exclusively to update $V_{\text{style}}^{(i)}$. Upon completion, this stage yields a set of style vectors $\{V_{\text{style}}^{(i)}\}$, where each vector encodes the fine-grained style of its corresponding image within the SD model's text embedding space.

\section{Stage 2: Pre-training the Style Encoder and Projection Layer}
\label{sec:stage2}
Stage 1 provides a unique style vector for each image but requires a time-consuming iterative optimization for any new image. To overcome this, Stage 2 introduces a feed-forward approach by training a style module, consisting of a style encoder $E_{\text{style}}$ and a style projection layer $P$, to predict the style vector from an image in a single pass.

The style encoder $E_{\text{style}}$ requires robust visual feature extraction capabilities. To ensure compatibility with Stable Diffusion, we instantiate $E_{\text{style}}$ with the pre-trained CLIP vision encoder (specifically, `openai/clip-vit-large-patch14`) that corresponds to SD v1.4's text encoder, leveraging its powerful, general-purpose visual understanding. The encoder takes a normalized $224 \times 224$ image $I$ and outputs a style feature $f_{\text{style}} = E_{\text{style}}(I)$, which is the `[CLS]` token's feature vector of dimension $d_{\text{enc}} = 768$.

The style projection layer $P$ is responsible for mapping the extracted style feature $f_{\text{style}}$ into the 8-token style vector space defined in Stage 1. Thus, $P: \mathbb{R}^{d_{\text{enc}}} \to \mathbb{R}^{8 \times d_{\text{text}}}$. We implement $P$ as a linear layer.

The pre-training in this stage consists of two steps. First, we pre-train only the style encoder $E_{\text{style}}$ using the image-tag pairs $(I, Tag)$. The objective is to teach $E_{\text{style}}$ to extract features that are discriminative of broad style categories. We use a cosine similarity loss $\mathcal{L}_{\text{clip}}$ to maximize the similarity between the image feature $f_{\text{style}}^{(i)} = E_{\text{style}}(I_i)$ and the text feature of its corresponding style tag $E_{\text{text}}(\text{Tag}_i)$:
\begin{equation}
\label{eq:clip_loss}
\mathcal{L}_{\text{clip}} = \mathbb{E}_{(I_i, \text{Tag}_i)} \left[ 1 - \frac{E_{\text{style}}(I_i) \cdot E_{\text{text}}(\text{Tag}_i)}{\|E_{\text{style}}(I_i)\|_2 \|E_{\text{text}}(\text{Tag}_i)\|_2} \right]
\end{equation}
After this, we freeze the encoder's parameters and pre-train only the projection layer $P$. The goal is to teach $P$ to map the extracted style feature $f_{\text{style}}^{(i)}$ to its corresponding target style vector $V_{\text{style}}^{(i)}$ learned in Stage 1. We use an MSE loss for this mapping objective:
\begin{equation}
\label{eq:map_loss}
\mathcal{L}_{\text{map}} = \mathbb{E}_{(I_i, V_{\text{style}}^{(i)})} \left\| P(E_{\text{style}}(I_i)) - V_{\text{style}}^{(i)} \right\|_2^2
\end{equation}
Upon completion, the combined module $P(E_{\text{style}}(I))$ can efficiently predict a style vector for any input image, providing a generalizable alternative to the per-image optimization of Stage 1.

\section{Stage 3: Joint Fine-tuning of the Style Module}
\label{sec:stage3}
While Stage 2 provides an efficient style extraction module, its training is indirect—it learns to mimic the outputs of Stage 1 rather than being optimized directly for the final task of image generation. A gap may exist between this imitation objective and the true objective of guiding the SD U-Net.

To bridge this gap, Stage 3 performs an end-to-end, joint fine-tuning of the entire style module. We unfreeze the parameters of both the style encoder $E_{\text{style}}$ and the projection layer $P$, allowing them to be updated simultaneously. The training process uses the image-caption pairs $(I, C)$ from our training set and optimizes the module directly with the SD reconstruction loss. This forces the encoder to learn features most salient for stylistic reconstruction and ensures the projection layer translates these features into the most effective conditioning for the U-Net.

The training flow is similar to Stage 1, but the conditioning vector $Cond'$ is now generated dynamically. For each sample $(I, C)$:
\begin{enumerate}
    \item The image $I$ is passed through the style module to get a predicted style vector: $V'_{\text{style}} = P(E_{\text{style}}(I))$.
    \item This predicted style vector $V'_{\text{style}}$ is concatenated with the text embedding of the content caption $E_{\text{text}}(C)$.
    \item This combined embedding serves as the condition $Cond'$ for the frozen SD U-Net.
\end{enumerate}
The fine-tuning loss $\mathcal{L}_{\text{finetune}}$ minimizes the noise prediction error, and its gradient is backpropagated to update the parameters of both $E_{\text{style}}$ and $P$:
\begin{equation}
\label{eq:finetune_loss}
\mathcal{L}_{\text{finetune}} = \mathbb{E}_{I, C, t, \epsilon} \left\| \epsilon - \epsilon_{\theta}(z_t, t, Cond') \right\|_2^2
\end{equation}
where $Cond' = [P(E_{\text{style}}(I)); E_{\text{text}}(C)]$. This end-to-end optimization yields the final style encoder $E_{\text{style}}^*$ and projection layer $P^*$, which are highly compatible with the downstream generator and possess superior instruction-following capabilities.

\section{Inference Pipeline}
\label{sec:inference}
During inference, our system takes two inputs from the user: a style reference image $I_{\text{ref}}$ and a new textual content description $C_{\text{new}}$. The goal is to generate an output image $I_{\text{out}}$ that inherits the style of $I_{\text{ref}}$ and depicts the content of $C_{\text{new}}$.

The process begins by feeding the preprocessed $I_{\text{ref}}$ through our trained style module to obtain the 8-token style vector: $V_{\text{style}} = P^*(E_{\text{style}}^*(I_{\text{ref}}))$. This vector encapsulates the fine-grained style information. Simultaneously, the content prompt $C_{\text{new}}$ is encoded into its text embeddings. The final conditioning $Cond_{\text{final}}$ is formed by concatenating the style vector and the content embeddings: $Cond_{\text{final}} = [V_{\text{style}}; E_{\text{text}}(C_{\text{new}})]$.

This combined condition guides the standard reverse diffusion process. We employ Classifier-Free Guidance (CFG) to enhance prompt adherence and image quality. At each timestep $t$, the final noise prediction $\epsilon_{\text{final}}$ is a linear combination of a conditional prediction $\epsilon_{\text{cond}} = \epsilon_{\theta}(z_t, t, Cond_{\text{final}})$ and an unconditional prediction $\epsilon_{\text{uncond}}$ (using an empty string "" as the condition):
$$
\epsilon_{\text{final}} = \epsilon_{\text{uncond}} + s_{\text{cfg}} \cdot (\epsilon_{\text{cond}} - \epsilon_{\text{uncond}})
$$
where the guidance scale $s_{\text{cfg}}$ is a hyperparameter, set to $7.5$ in our experiments.

For the reverse diffusion sampling, we use the PNDM (Pseudo Numerical Methods for Diffusion Models) sampler. The latent variable $z_{t-1}$ is computed iteratively from $z_t$ for a total of $T_{\text{infer}}$ steps (typically 50). The update rule is given by:
\begin{equation}
z_{t-1} = \sqrt{\alpha_{t-1}} \left( \frac{z_t - \sqrt{1 - \alpha_t} \epsilon_{\text{final}}}{\sqrt{\alpha_t}} \right) + \sqrt{1 - \alpha_{t-1}} \epsilon_{\text{final}}
\end{equation}
where $\alpha_t$ are the diffusion schedule coefficients. Once the iterative process is complete, the final latent variable $z_0$ is decoded back to the pixel space using the frozen VAE decoder, yielding the final stylized image: $I_{\text{out}} = D_{\text{VAE}}(z_0)$.

\section{Experiments}
\label{sec:experiments}

This chapter details the phased experimental setup, the evaluation of results from each stage, and an ablation study to validate our architectural choices.

\subsection{Implementation Details}
This section outlines the experimental specifics, from dataset partitioning to the hyperparameter settings for our three-stage training process.

\subsubsection{Dataset Partitioning}
We partitioned the \texttt{Style30k-captions} dataset, consisting of $(I, C, Tag)$ triplets, into a training set and a test set. The test set contains 2,500 samples and is constructed to include examples from every style tag ($Tag$) present in the original dataset, ensuring comprehensive evaluation.

\subsubsection{Stage 1: Style Vector Inversion}
For each image $I_i$ in both the training and test sets, we initialized a style vector $V_{\text{style}}^{(i)} \in \mathbb{R}^{8 \times d_{\text{text}}}$ by sampling from $\mathcal{N}(0, 0.02)$. We used the AdamW optimizer with a learning rate of $5 \times 10^{-4}$ and a numerical stability constant $\epsilon = 1 \times 10^{-8}$. Each style vector was trained for 250 steps, with checkpoints saved every 50 steps for evaluation.

\subsubsection{Stage 2: Pre-training Style Encoder and Projection Layer}
In the style encoder pre-training phase, we initialized the encoder with weights from \texttt{openai/clip-vit-large-patch14} and trained it on the $(I_{\text{train}}, Tag_{\text{train}})$ pairs from the \texttt{Style30k-captions} training set. We used the AdamW optimizer with a batch size of 32, a learning rate of $5 \times 10^{-5}$, $\epsilon = 1 \times 10^{-8}$, and trained for 10 epochs.

For the style projection layer pre-training, we froze the style encoder's weights and trained only the projection layer's parameters. This was done using the training set images $I_{\text{train}}$ and their corresponding style vectors $V_{\text{style}}^{(i)}$ obtained from Stage 1. We again used the AdamW optimizer with a batch size of 32, a learning rate of $5 \times 10^{-5}$, $\epsilon = 1 \times 10^{-8}$, and trained for 5 epochs.

\subsubsection{Stage 3: Joint Fine-tuning of the Style Module}
To further improve the quality and controllability of the generated images, we unfroze both the style encoder and the projection layer for joint fine-tuning. The training was performed on the $(I_{\text{train}}, C_{\text{train}})$ pairs. We used the AdamW optimizer with a batch size of 8 and trained for 10 epochs. The learning rate for the projection layer was fixed at $5 \times 10^{-4}$. For the style encoder, we experimented with three different learning rates: $1 \times 10^{-5}$, $5 \times 10^{-6}$, and $2 \times 10^{-6}$.

\subsection{Qualitative and Quantitative Evaluation}
This section presents the evaluation of our three-stage training process. We assess the results from three perspectives: visual quality, style similarity, and image-text alignment.

\subsubsection{Evaluation of Stage 1 Style Vectors}
To determine the optimal number of training steps for the textual inversion process, we evaluated the quality of the learned style vectors at different checkpoints. We used the pre-trained style encoder from Stage 2 (trained only with style tags) as a feature extractor to compute a style similarity score. Specifically, for each original image $I_i$, we generated a reconstructed image $I_i'$ using its content caption and its style vector from a given step. The style score is the cosine similarity between their features:
\begin{equation}
\text{StyleScore}_i = \frac{E(I_i') \cdot E(I_i)}{\|E(I_i')\|_2 \|E(I_i)\|_2}
\end{equation}
where $E$ is the pre-trained style encoder. Additionally, to assess text-to-image alignment, we calculated the standard CLIP score (cosine similarity between the image features of $I_i'$ and the text features of the prompt). The averaged results are shown in Table \ref{tab:stage1_eval}.

\begin{table}[h]
\centering
\caption{Style and image-text similarity at different training steps for Stage 1. Best scores are in bold.}
\label{tab:stage1_eval}
\begin{tabular}{l c c c c}
\toprule
\textbf{Metric} & \textbf{Step 50} & \textbf{Step 100} & \textbf{Step 150} & \textbf{Step 200} \\
\midrule
\textbf{Style Similarity} & 0.9118 & 0.9164 & 0.9152 & \textbf{0.9226} \\
\textbf{Image-Text Sim.} & 0.2730 & 0.2732 & \textbf{0.2754} & 0.2748 \\
\bottomrule
\end{tabular}
\end{table}

\begin{figure*}[!t]
\centering
\begin{tabular}{c c}
    \textbf{EX.1} & \includegraphics[width=0.85\textwidth]{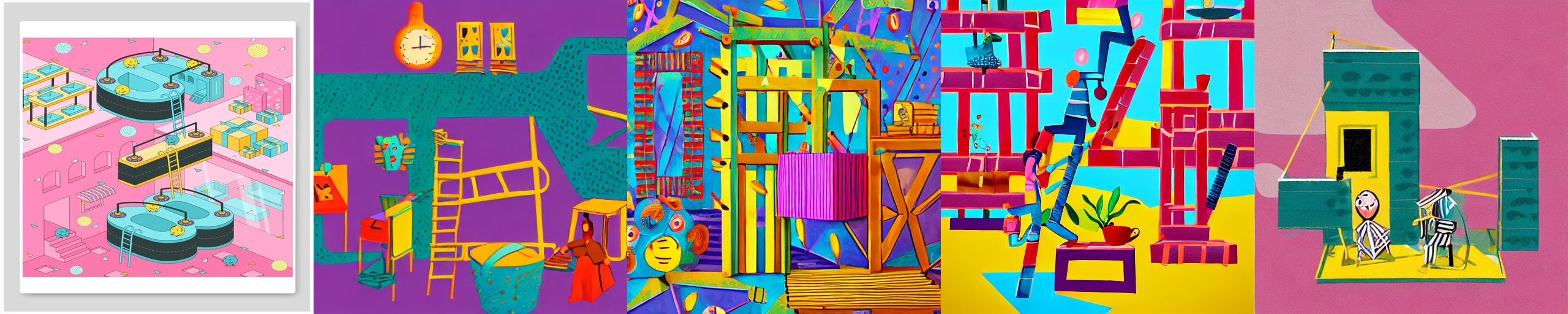} \\
    \textbf{EX.2} & \includegraphics[width=0.85\textwidth]{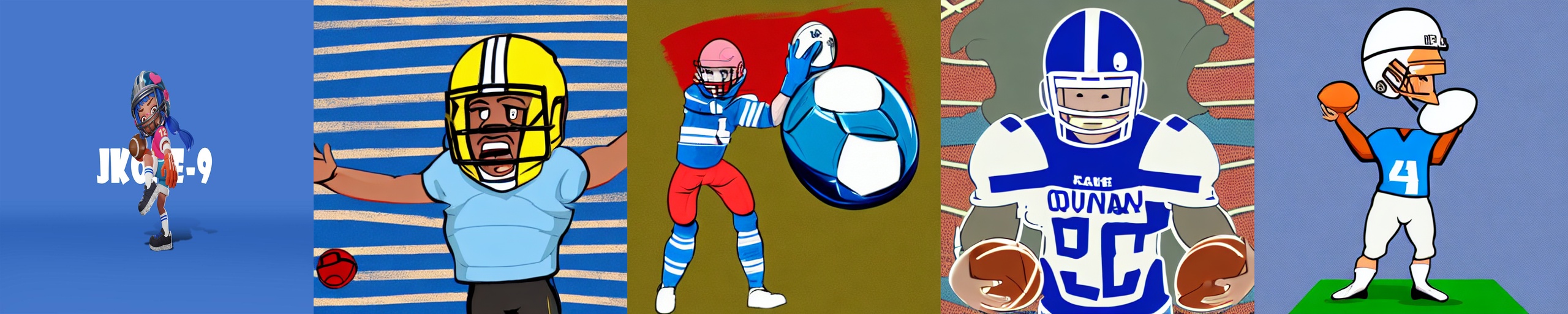} \\
    \textbf{EX.3} & \includegraphics[width=0.85\textwidth]{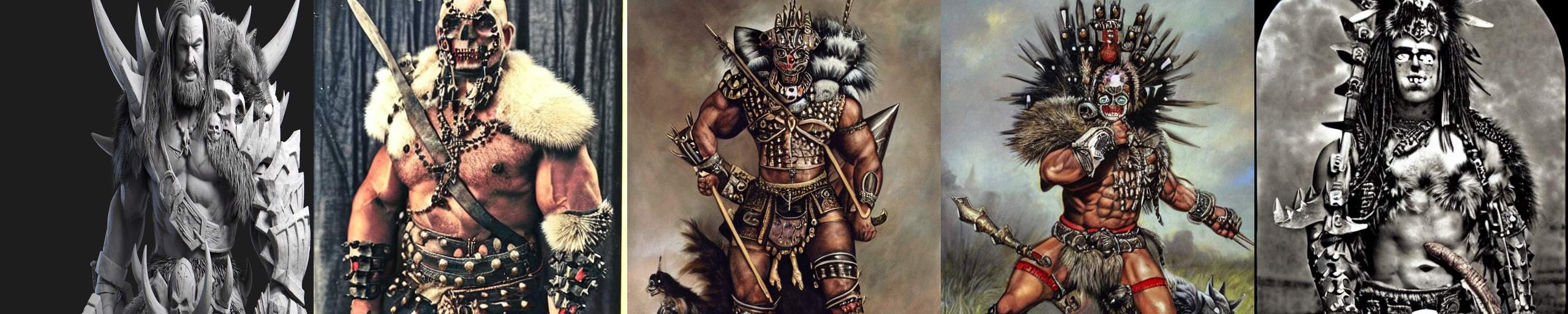} \\
    \textbf{EX.4} & \includegraphics[width=0.85\textwidth]{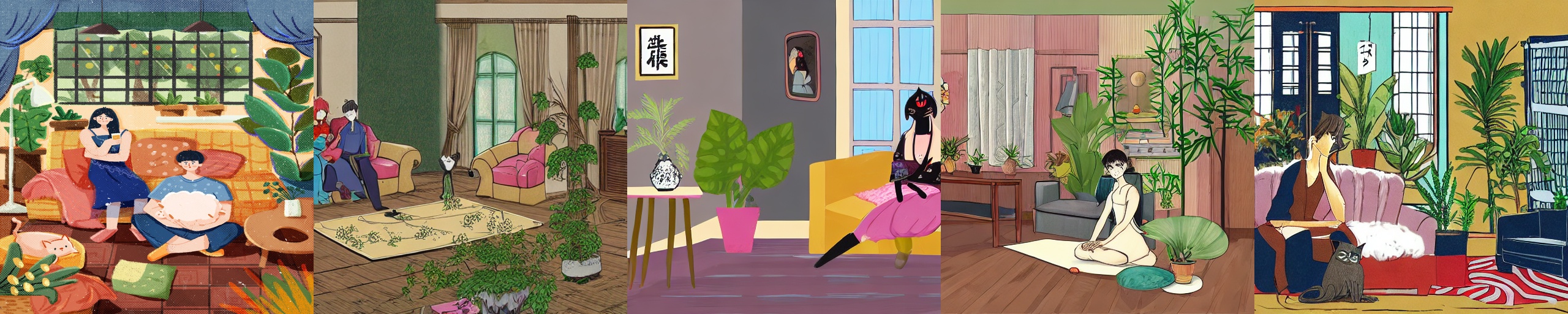} \\
    \textbf{EX.5} & \includegraphics[width=0.85\textwidth]{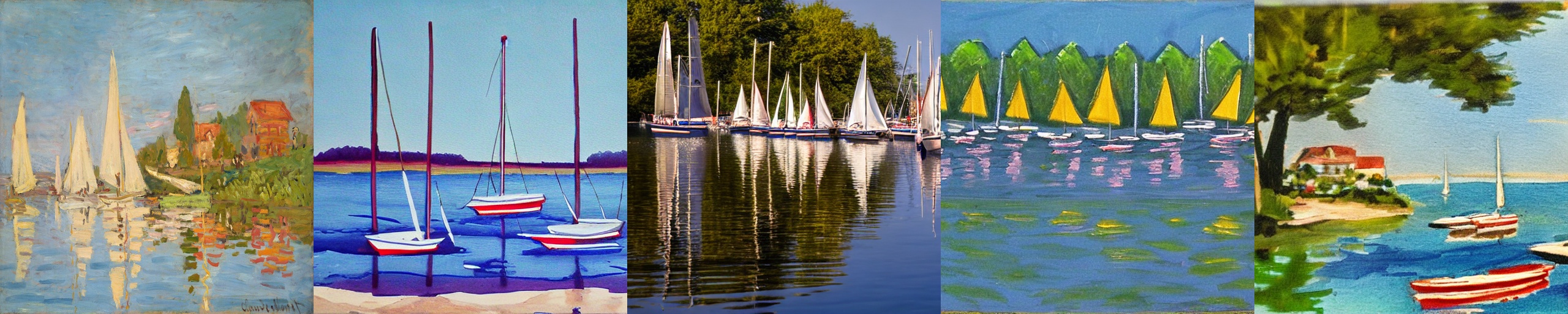} \\
    \textbf{EX.6} & \includegraphics[width=0.85\textwidth]{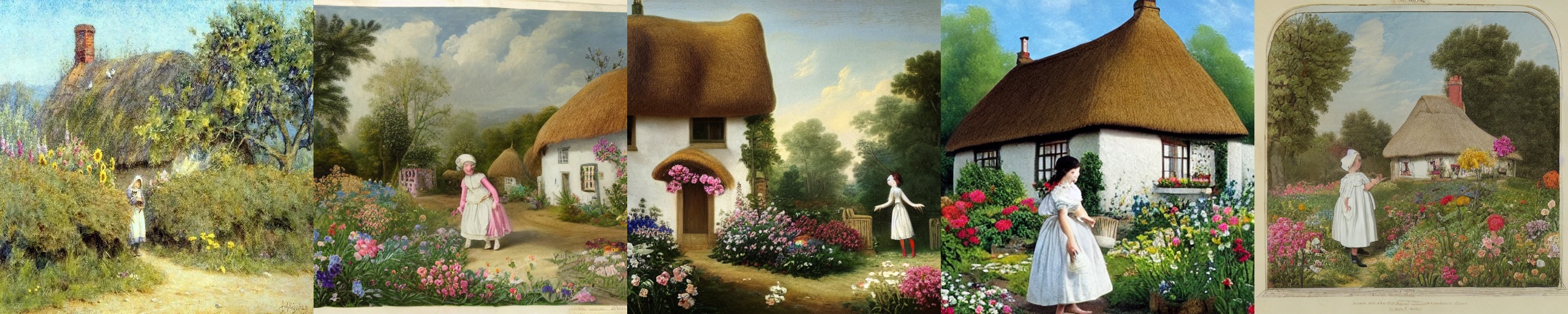} \\
\end{tabular}
\caption{Visual evaluation of reconstruction quality at different training steps in Stage 1. Each row shows one example. The images within each row, from left to right, are: Original Image, Reconstruction at Step 50, Step 100, Step 150, and Step 200. In most cases, the Step 200 reconstruction is the most faithful.}
\label{fig:stage1_recon_vis}
\end{figure*}

Based on the highest style similarity for reconstructions, we selected the style vectors trained for 200 steps as the targets for the subsequent training of the style projection layer in Stage 2. Visual results in Figure \ref{fig:stage1_recon_vis} confirm that reconstruction quality improves with more training steps, with step 200 generally yielding the best results.

\subsubsection{Evaluation of Style Encoder Pre-training}
To evaluate the feature extraction capability of our pre-trained style encoder, we compare it against the original CLIP vision encoder and a pre-trained VGG19 model. We randomly selected 10 style classes and extracted features for all images within them using each method. For VGG19, we extracted Gram matrices from three layers to represent style. We then used t-SNE to visualize the high-dimensional features in a 2D plane.

\begin{figure}[!hbtp]
\centering
\includegraphics[width=\columnwidth]{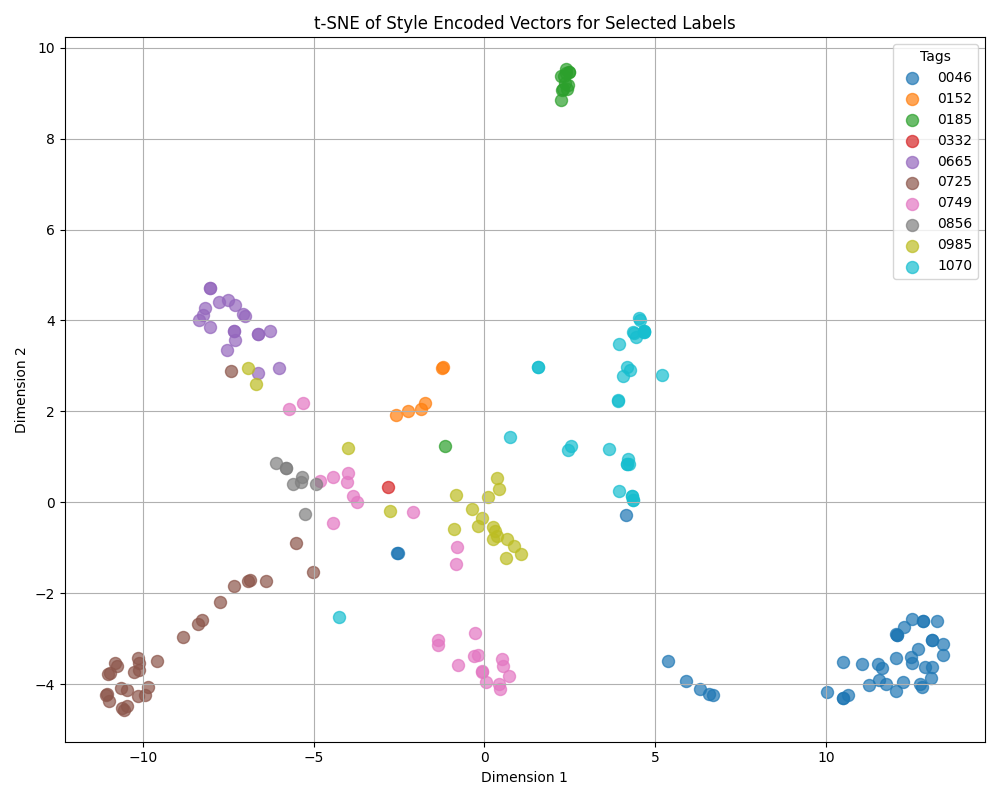}
\vspace{2mm}
\includegraphics[width=\columnwidth]{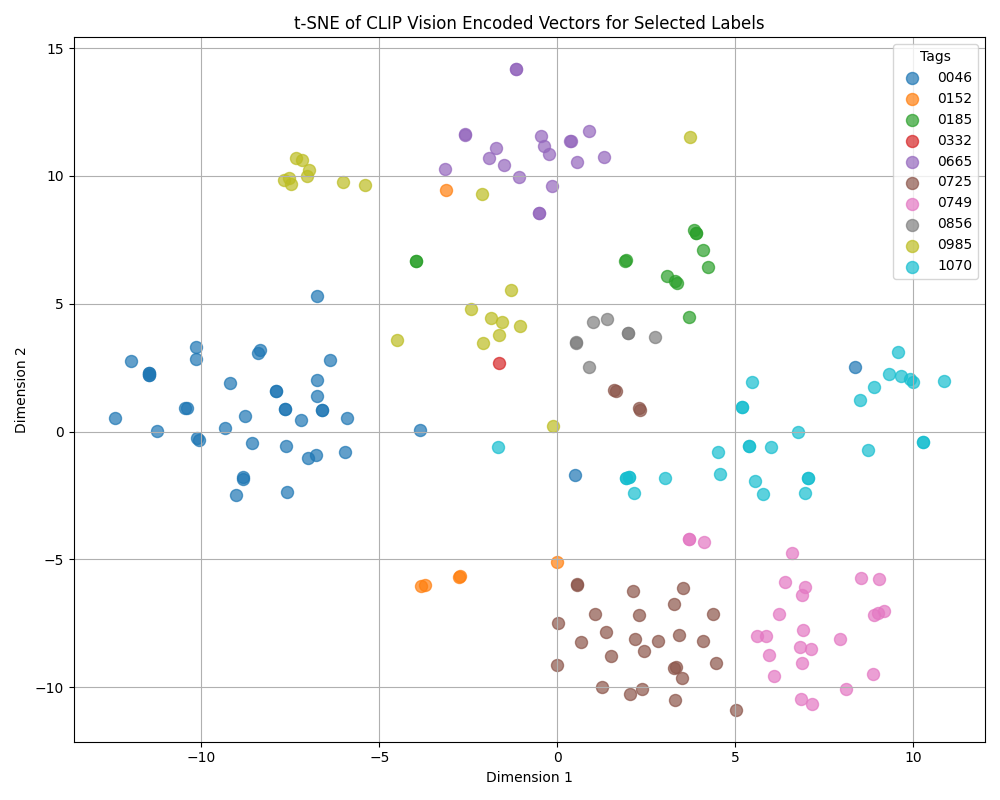}
\vspace{2mm}
\includegraphics[width=\columnwidth]{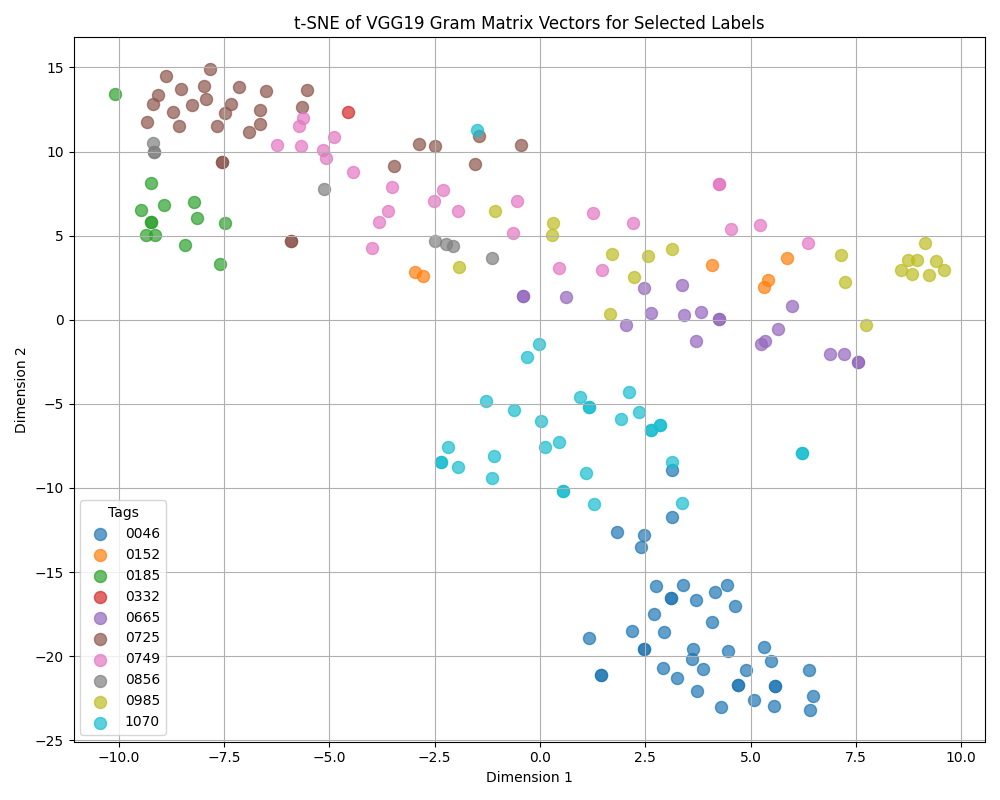}

\caption{t-SNE visualization of features extracted by different encoders. Each color represents a different style category. Our pre-trained style encoder (a) demonstrates superior clustering of style categories compared to both the baseline CLIP (b) and VGG19 (c) models.}
\label{fig:tsne_vis}
\end{figure}

As shown in Figure \ref{fig:tsne_vis}, the data points for our style encoder form more distinct and coherent clusters for different style categories compared to the other two models, indicating its enhanced ability to perceive and group images by style.

\subsubsection{Evaluation of Stage 2 Pre-trained Module}
After pre-training, the style encoder and projection layer can be used to directly infer a style vector from an image without iterative optimization. We evaluated the quality of this feed-forward approach on the test set using the same metrics as in Stage 1. Table \ref{tab:stage1_vs_stage2_eval} compares the performance of the Stage 2 module against the per-image optimized vectors from Stage 1.

\begin{table}[h]
\centering
\caption{Comparison of reconstruction quality metrics between Stage 1 (per-image inversion) and Stage 2 (feed-forward inference) on the test set.}
\label{tab:stage1_vs_stage2_eval}
\begin{tabular}{l c c}
\toprule
\textbf{Style Vector Source} & \textbf{Stage 1 (200 steps)} & \textbf{Stage 2 (Inference)} \\
\midrule
\textbf{Style Similarity} & \textbf{0.9226} & 0.9208 \\
\textbf{Image-Text Sim.} & \textbf{0.2748} & 0.2740 \\
\bottomrule
\end{tabular}
\end{table}

\begin{figure*}[!t]
\centering
\begin{tabular}{cccc}
     \multicolumn{2}{c}{\textbf{Example 1}} & \multicolumn{2}{c}{\textbf{Example 2}} \\
     \includegraphics[width=0.2\textwidth]{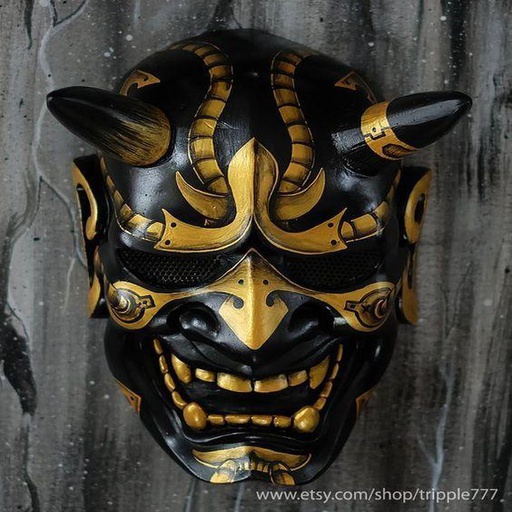} & \includegraphics[width=0.2\textwidth]{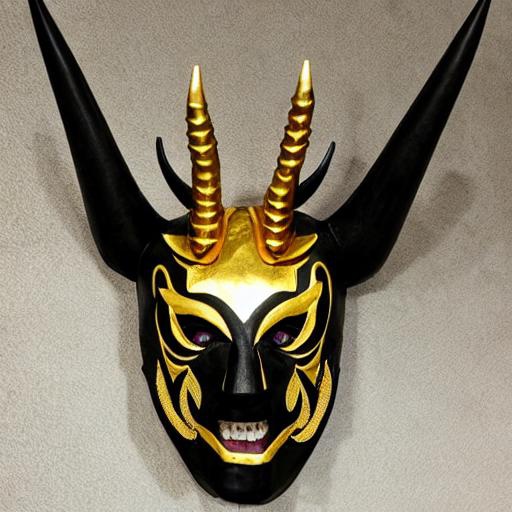} &
     \includegraphics[width=0.2\textwidth]{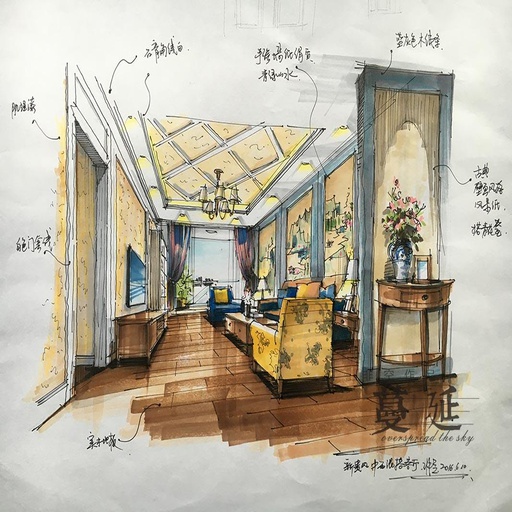} & \includegraphics[width=0.2\textwidth]{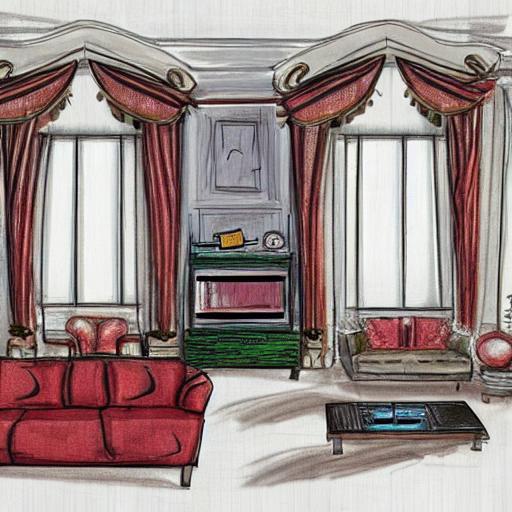} \\
     (Original) & (Stage 2 Recon.) & (Original) & (Stage 2 Recon.) \\[4mm]
     
     \multicolumn{2}{c}{\textbf{Example 3}} & \multicolumn{2}{c}{\textbf{Example 4}} \\
     \includegraphics[width=0.2\textwidth]{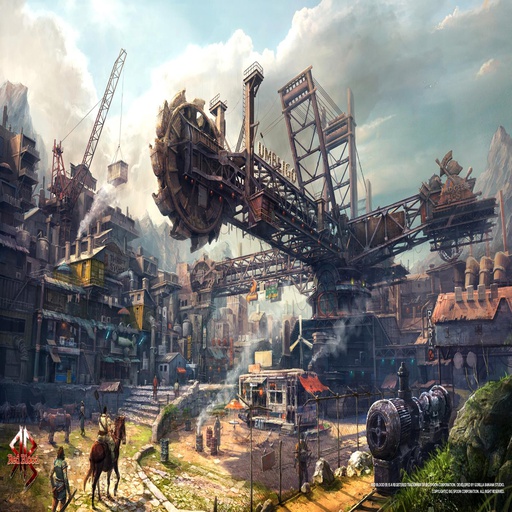} & \includegraphics[width=0.2\textwidth]{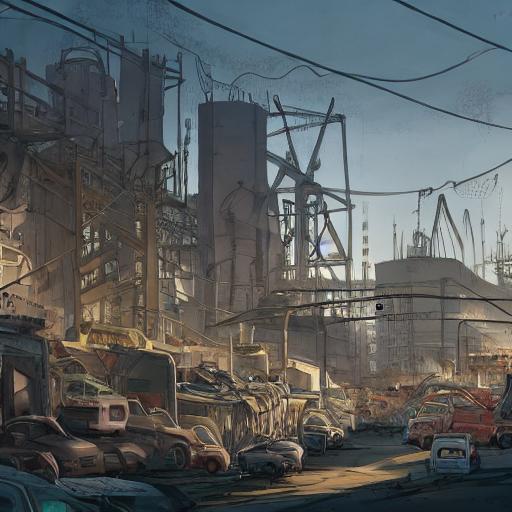} &
     \includegraphics[width=0.2\textwidth]{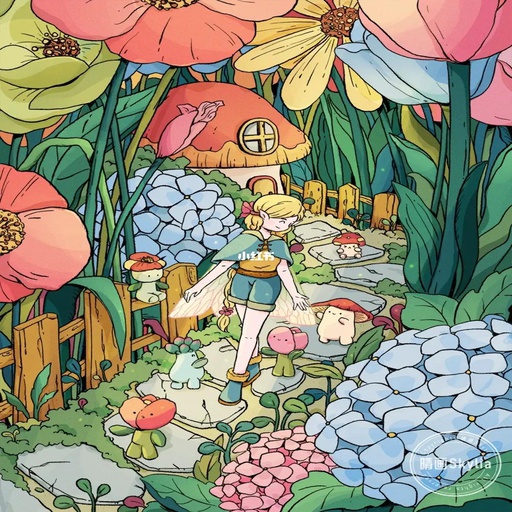} & \includegraphics[width=0.2\textwidth]{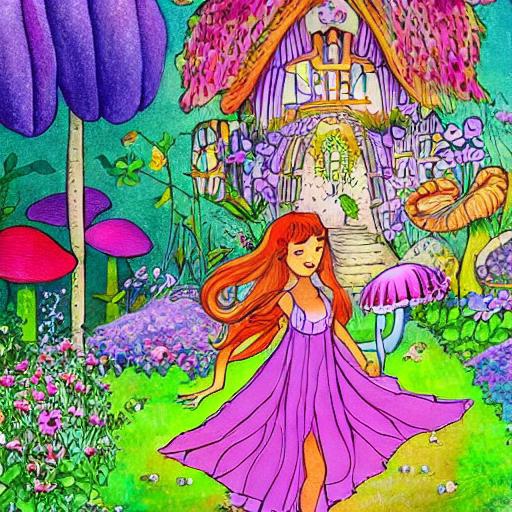} \\
     (Original) & (Stage 2 Recon.) & (Original) & (Stage 2 Recon.) \\[4mm]

     \multicolumn{2}{c}{\textbf{Example 5}} & \multicolumn{2}{c}{\textbf{Example 6}} \\
     \includegraphics[width=0.2\textwidth]{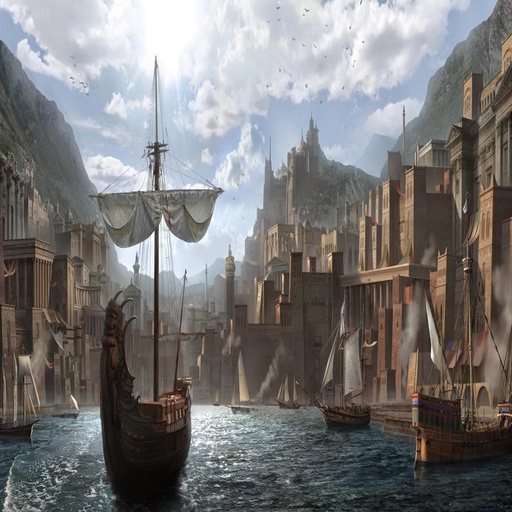} & \includegraphics[width=0.2\textwidth]{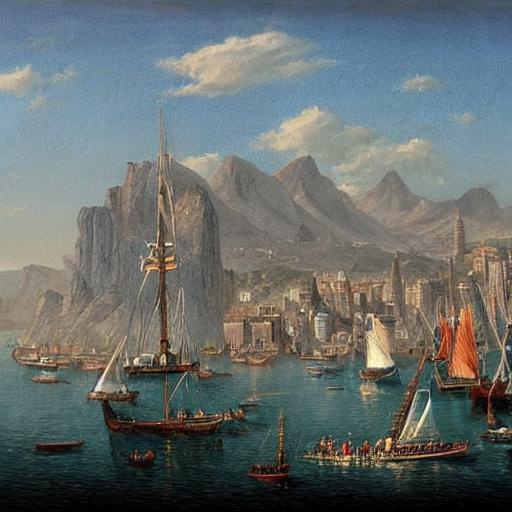} &
     \includegraphics[width=0.2\textwidth]{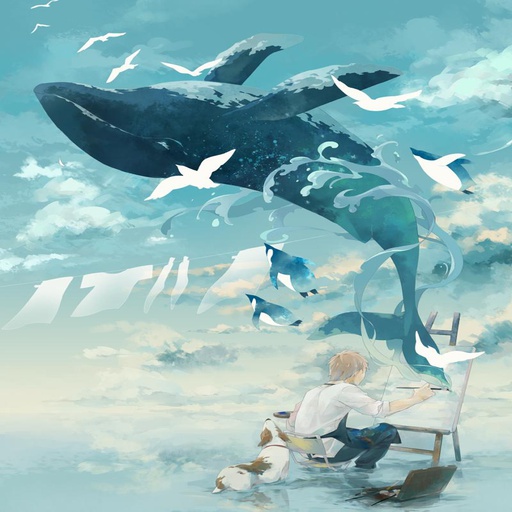} & \includegraphics[width=0.2\textwidth]{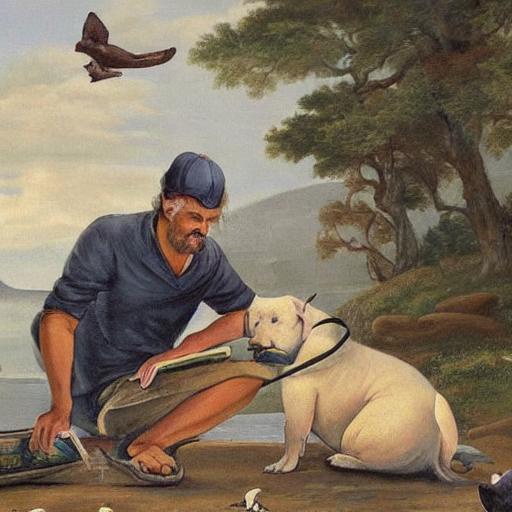} \\
     (Original) & (Stage 2 Recon.) & (Original) & (Stage 2 Recon.) \\
\end{tabular}
\caption{Visual comparison of original images and their reconstructions using the Stage 2 module. In each example, the left image is the original and the right is the reconstruction. The results show high fidelity.}
\label{fig:compare-stage2}
\end{figure*}

The results show that the Stage 2 module achieves performance nearly on par with the computationally expensive per-image inversion from Stage 1. This demonstrates that our pre-trained module successfully generalizes to unseen test data, providing a highly efficient yet effective paradigm for style vector extraction. Visual comparisons are provided in Figure \ref{fig:compare-stage2}.

\subsubsection{Evaluation of Stage 3 Joint Fine-tuning}
Finally, we evaluated the fully fine-tuned module from Stage 3. We tested three different learning rates (LR) for the style encoder: Large ($1 \times 10^{-5}$), Medium ($5 \times 10^{-6}$), and Small ($2 \times 10^{-6}$).

\begin{table}[h]
\centering
\caption{Final comparison of all three stages on the test set. LR refers to the style encoder learning rate in Stage 3.}
\label{tab:stage3_eval}
\begin{tabular}{l c c c c c}
\toprule
\textbf{Metric} & \textbf{Stage 1} & \textbf{Stage 2} & \textbf{S3 (LR-L)} & \textbf{S3 (LR-M)} & \textbf{S3 (LR-S)} \\
\midrule
\textbf{Style Sim.} & 0.9226 & 0.9208 & 0.9215 & 0.9229 & \textbf{0.9305} \\
\textbf{Img-Txt Sim.} & \textbf{0.2748} & 0.2740 & 0.2712 & 0.2731 & 0.2744 \\
\bottomrule
\end{tabular}
\end{table}

\begin{figure*}[!t]
\centering
\begin{tabular}{cccc}
     \multicolumn{2}{c}{\textbf{Example 1}} & \multicolumn{2}{c}{\textbf{Example 2}} \\
     \includegraphics[width=0.2\textwidth]{newimage/newimages/01.jpg} & \includegraphics[width=0.2\textwidth]{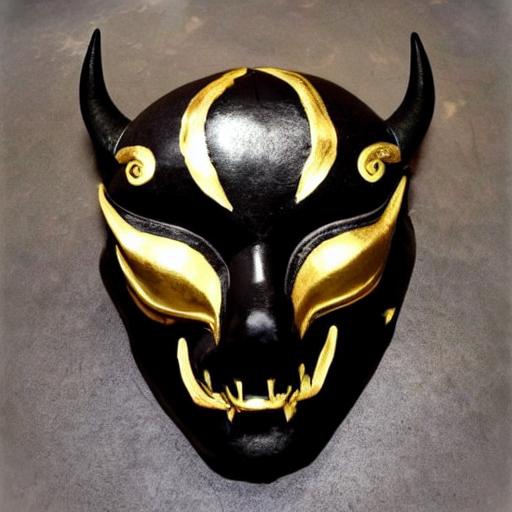} &
     \includegraphics[width=0.2\textwidth]{newimage/newimages/02.jpg} & \includegraphics[width=0.2\textwidth]{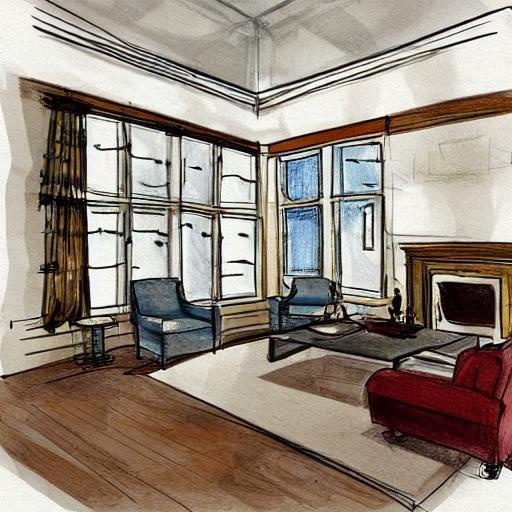} \\
     (Original) & (Stage 3 Recon.) & (Original) & (Stage 3 Recon.) \\[4mm]
     
     \multicolumn{2}{c}{\textbf{Example 3}} & \multicolumn{2}{c}{\textbf{Example 4}} \\
     \includegraphics[width=0.2\textwidth]{newimage/newimages/05.jpg} & \includegraphics[width=0.2\textwidth]{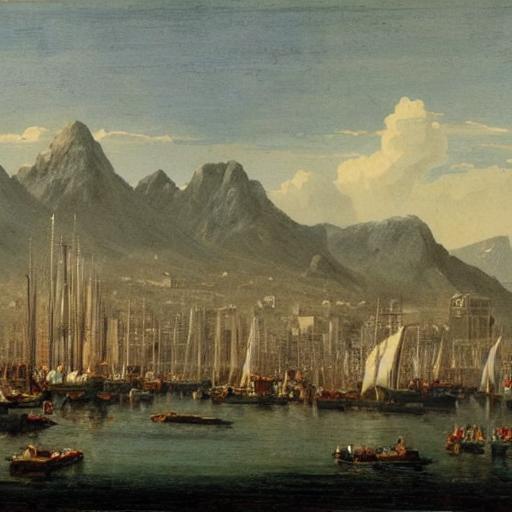} &
     \includegraphics[width=0.2\textwidth]{newimage/newimages/06.jpg} & \includegraphics[width=0.2\textwidth]{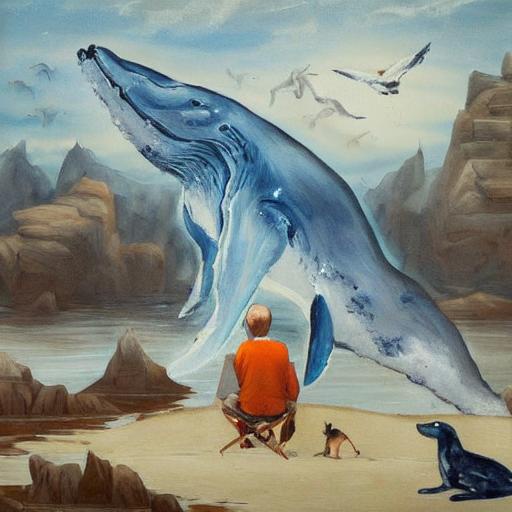} \\
     (Original) & (Stage 3 Recon.) & (Original) & (Stage 3 Recon.) \\[4mm]

     \multicolumn{2}{c}{\textbf{Example 5}} & \multicolumn{2}{c}{\textbf{Example 6}} \\
     \includegraphics[width=0.2\textwidth]{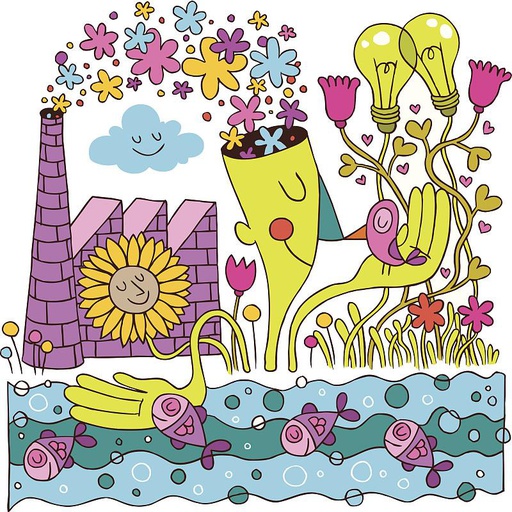} & \includegraphics[width=0.2\textwidth]{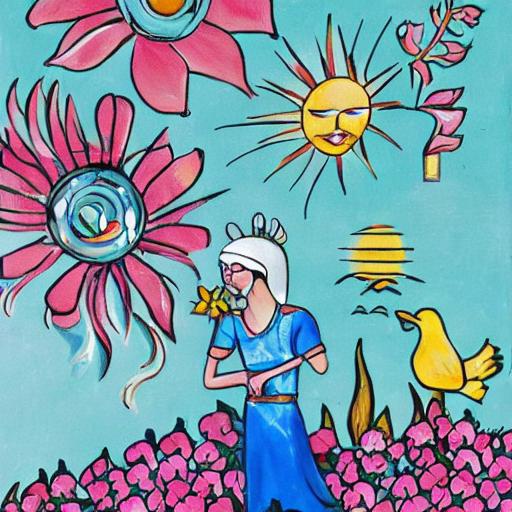} &
     \includegraphics[width=0.2\textwidth]{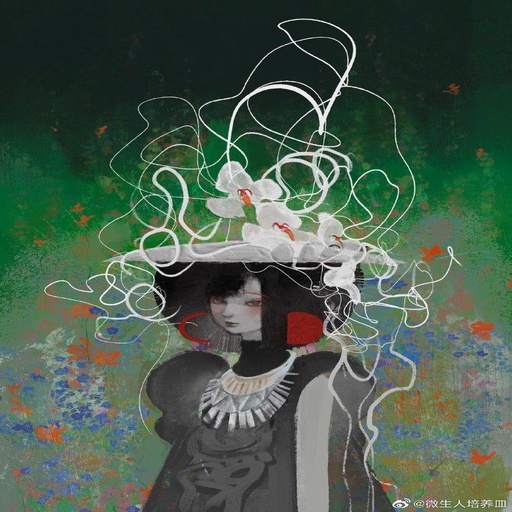} & \includegraphics[width=0.2\textwidth]{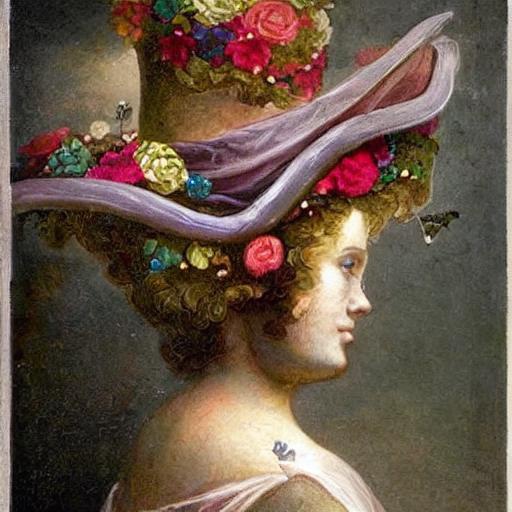} \\
     (Original) & (Stage 3 Recon.) & (Original) & (Stage 3 Recon.) \\
\end{tabular}
\caption{Visual comparison of original images and their reconstructions using the fully fine-tuned Stage 3 module. In each example, the left image is the original and the right is the reconstruction. Stage 3 shows a noticeable improvement in capturing fine-grained stylistic details over Stage 2.}
\label{fig:compare-stage3}
\end{figure*}

As shown in Table \ref{tab:stage3_eval}, the joint fine-tuning in Stage 3, particularly with a small learning rate for the encoder, yields the best style similarity, surpassing even the per-image optimization of Stage 1. While the image-text similarity sees a marginal dip, the significant gain in style fidelity indicates that the end-to-end optimization successfully tunes the module for superior style control. Figure \ref{fig:compare-stage3} shows the qualitative improvement.

\subsection{Ablation Study}
\begin{figure*}[!t]
\centering
\begin{tabular}{cccc}
     \multicolumn{2}{c}{\textbf{Example 1}} & \multicolumn{2}{c}{\textbf{Example 2}} \\
     \includegraphics[width=0.2\textwidth]{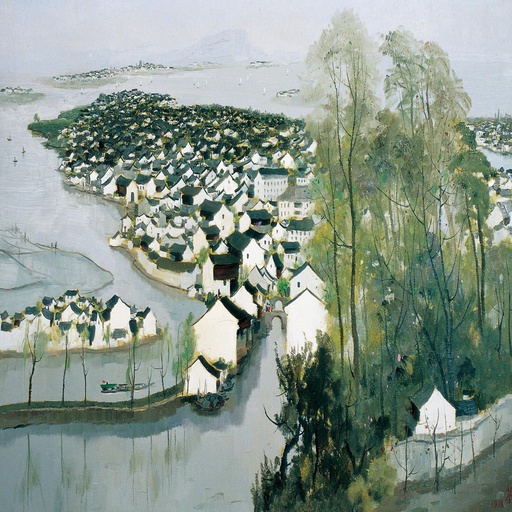} & \includegraphics[width=0.2\textwidth]{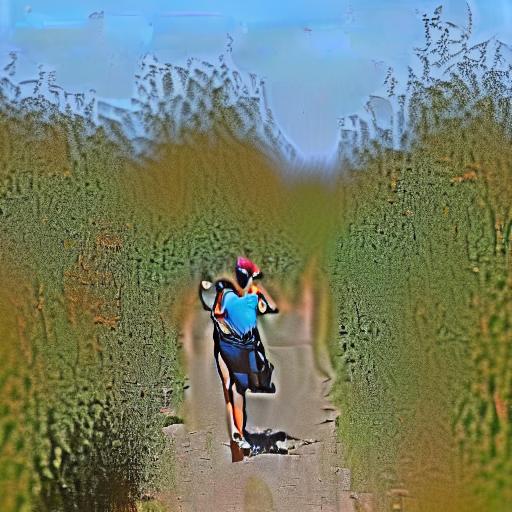} &
     \includegraphics[width=0.2\textwidth]{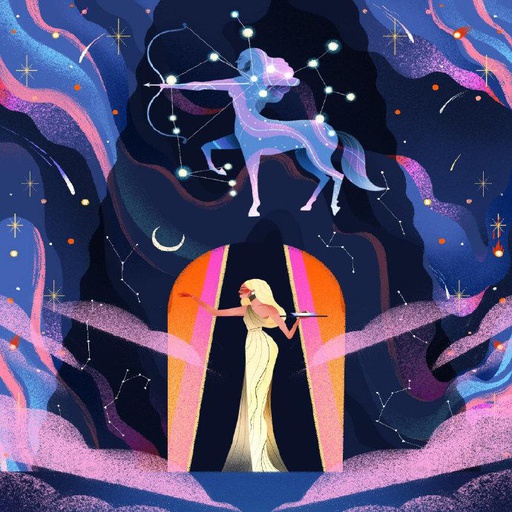} & \includegraphics[width=0.2\textwidth]{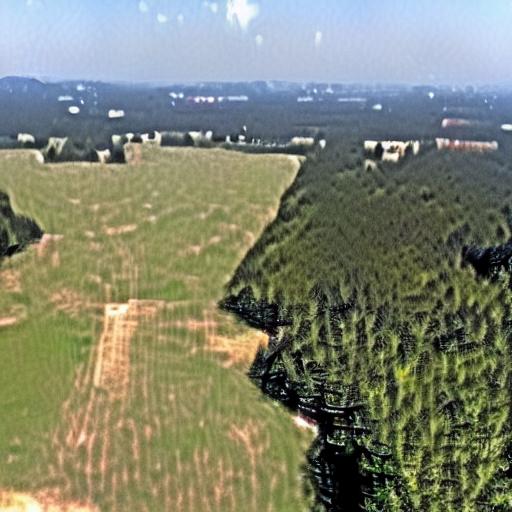} \\
     (Original) & (Ablation Recon.) & (Original) & (Ablation Recon.) \\[4mm]
     
     \multicolumn{2}{c}{\textbf{Example 3}} & \multicolumn{2}{c}{\textbf{Example 4}} \\
     \includegraphics[width=0.2\textwidth]{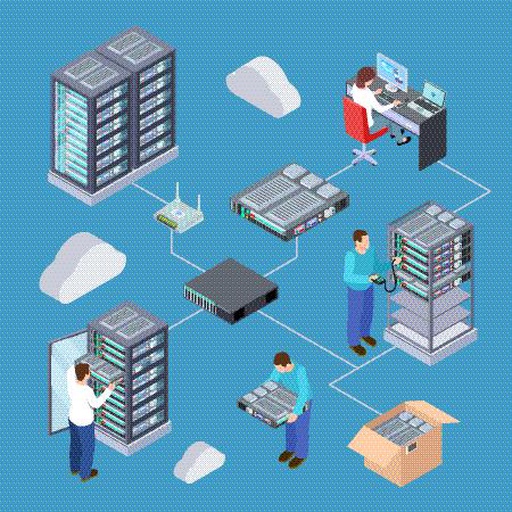} & \includegraphics[width=0.2\textwidth]{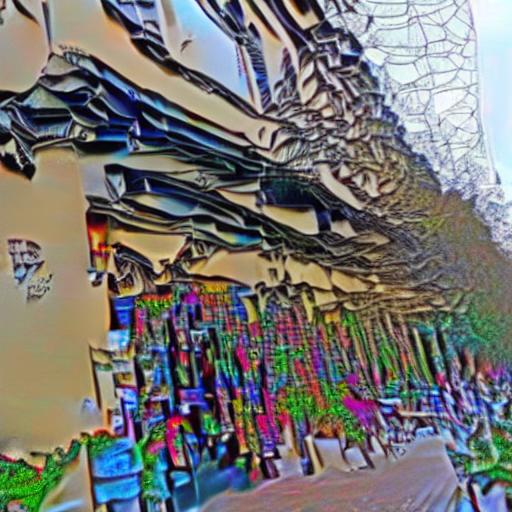} &
     \includegraphics[width=0.2\textwidth]{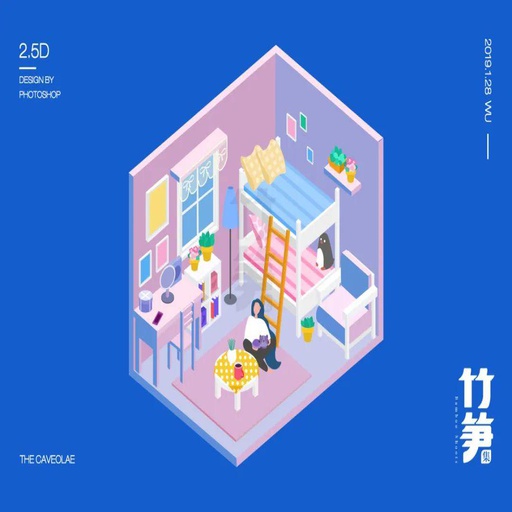} & \includegraphics[width=0.2\textwidth]{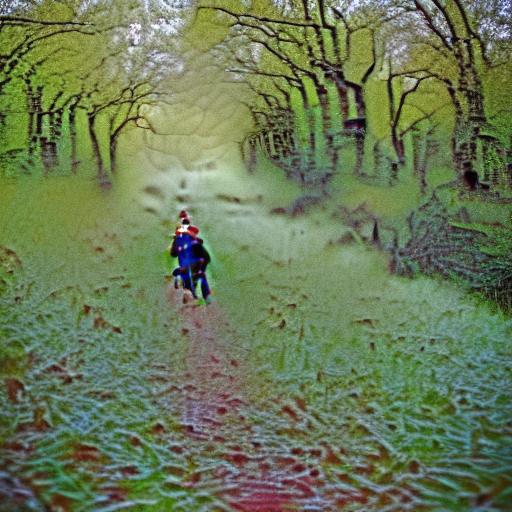} \\
     (Original) & (Ablation Recon.) & (Original) & (Ablation Recon.) \\
\end{tabular}
\caption{Reconstruction results from the ablation study without the projection layer. In each example, the left image is the original and the right is the reconstruction. The model fails to generate semantically coherent images.}
\label{fig:ablation_results}
\end{figure*}

To validate the necessity of the style projection layer ($P$), we conducted an ablation study. In this experiment, we removed the projection layer and instead fine-tuned the style encoder directly, using its output feature vector replicated 8 times to form the style tokens. We used the best-performing hyperparameters from Stage 3 (encoder LR of $2 \times 10^{-6}$) and trained for 3 epochs. The results in Figure \ref{fig:ablation_results} show that the model without the projection layer completely fails to generate meaningful images. This result strongly suggests that the learnable projection layer is a critical component, acting as an essential bridge to translate the visual features from the encoder into a format that the text-conditioned U-Net can effectively interpret.

\section{Conclusion}
\label{sec:conclusion}

This paper addressed the challenge of fine-grained style extraction and generation from few-shot image examples, aiming to overcome the limitations of existing text-to-image models when handling specific artistic styles that are difficult to describe with natural language. While current diffusion-based generative models have made significant strides in image quality and text adherence, they still struggle to generate images with specific, nuanced styles under zero-shot or few-shot conditions. To tackle these challenges, we proposed a novel three-stage training framework for fine-grained stylized image generation.

Our contribution began at the data level, where we constructed the \texttt{Style30k-captions} dataset by leveraging the GPT-4o model to generate content-only descriptions, effectively decoupling style and content information. The core of our method is a three-stage training pipeline. First, we employed textual inversion to learn an 8-token style vector for each image, aligned with the text embedding space of Stable Diffusion and optimized to disentangle style from the provided content captions. Second, to improve efficiency and generalizability, we pre-trained a style encoder and a projection layer, enabling direct, single-pass style vector extraction from any image. Our experiments showed this feed-forward module outperformed baseline methods in feature clustering and achieved performance comparable to the costly per-image inversion. The third and final stage involved an end-to-end joint fine-tuning of the style module, directly optimizing its parameters against the image reconstruction loss. This step proved crucial, yielding the highest style similarity scores and confirming the importance of the projection layer via ablation studies.

In summary, this research successfully establishes an effective framework capable of extracting fine-grained style attributes from a single reference image and generating controllable, high-quality stylized images guided by new text prompts. By integrating a multi-stage training process with multimodal alignment strategies, our method enhances the style control capabilities of diffusion models while preserving their powerful content generation abilities, paving the way for new applications in personalized visual content creation.

Looking forward, while this work presents a robust framework, several avenues for future research remain. The model's generalization could be enhanced by expanding its training to more diverse style domains, such as photographic styles, 3D renders, or specific designer aesthetics. There is also room to investigate more advanced disentanglement techniques, like adversarial learning or refined attention mechanisms, for an even cleaner separation of style and content. Furthermore, improving computational efficiency through knowledge distillation or advanced samplers would be beneficial for real-time and interactive applications. Finally, the modularity of our style extractor invites exploration into its integration with other generative modalities, such as video style transfer, and the development of more complex controls that allow users to creatively combine or modulate stylistic elements from multiple sources.

\end{document}